\documentclass[preprint,12pt]{elsarticle}

\makeatletter
\def\ps@pprintTitle{%
\let\@oddhead\@empty
\let\@evenhead\@empty
\def\@oddfoot{}%
\let\@evenfoot\@oddfoot}
\makeatother

\usepackage{amssymb}
\usepackage{amsmath}
\usepackage{url}
\usepackage{subcaption}
\usepackage{multirow}
\usepackage{hyphenat}
\usepackage{xparse}
\usepackage{color}
\usepackage{dsfont}
\begin{document}

\begin{frontmatter}



\title{FADE: Forecasting for Anomaly Detection on ECG} 


\author[1]{Paula Ruiz-Barroso\corref{cor}}
\ead{pruizb@uma.es}
\author[1]{Francisco M. Castro}
\ead{fcastro@uma.es}
\author[2]{José Miranda}
\ead{jose.mirandacalero@epfl.ch}
\author[2]{Denisa-Andreea Constantinescu}
\ead{denisa.constantinescu@epfl.ch}
\author[2]{David Atienza}
\ead{david.atienza@epfl.ch}
\author[1]{Nicol\'as Guil}
\ead{nguil@uma.es}

\cortext[cor]{Corresponding author. \textit{Postal address:} Bulevar Louis Pasteur, 35, Office 2.3.8a, 29071, Malaga, Spain. \textit{Phone number:} +34952133388}

\affiliation[1]{organization={Department of Computer Architecture, University of Málaga},
            city={Malaga},
            postcode={29071}, 
            country={Spain}}

\affiliation[2]{organization={Embedded Systems Laboratory at the École Polytechnique Fédérale of Lausanne},
            city={Lausanne},
            country={Switzerland}}

\begin{abstract}

\textbf{Background and Objective:} Cardiovascular diseases, a leading cause of noncommunicable disease-related deaths, require early and accurate detection to improve patient outcomes. Taking advantage of advances in machine learning and deep learning, multiple approaches have been proposed in the literature to address the challenge of detecting ECG anomalies. Typically, these methods are based on the manual interpretation of ECG signals, which is time consuming and depends on the expertise of healthcare professionals. The objective of this work is to propose a deep learning system, FADE, designed for normal ECG forecasting and anomaly detection, which reduces the need for extensive labeled datasets and manual interpretation.

\textbf{Methods:} We propose FADE, a deep learning system designed for normal ECG forecasting, trained in a self-supervised manner with a novel morphological inspired loss function, that can be used for anomaly detection. Unlike conventional models that learn from labeled anomalous ECG waveforms, our approach predicts the future of normal ECG signals, thus avoiding the need for extensive labeled datasets. Using a novel distance function to compare forecasted ECG signals with actual sensor data, our method effectively identifies cardiac anomalies. Additionally, this approach can be adapted to new contexts (e.g., different sensors, patients, etc.) through domain adaptation techniques. To evaluate our proposal, we performed a set of experiments using two publicly available datasets: MIT-BIH NSR and MIT-BIH Arrythmia.

\textbf{Results:} The results demonstrate that our system achieves an average accuracy of 83.84\% in anomaly detection, while correctly classifying normal ECG signals with an accuracy of 85.46\%.

\textbf{Conclusions:} Our proposed approach exhibited superior performance in the early detection of cardiac anomalies in ECG signals, surpassing previous methods that predominantly identify a limited range of anomalies. FADE effectively detects both abnormal heartbeats and arrhythmias, offering significant advantages in healthcare through cost reduction, facilitation of remote monitoring, and efficient processing of large-scale ECG data.

\end{abstract}




\end{frontmatter}



\section{Introduction}
\label{sec:introduction}

According to the World Health Organization (WHO), noncommunicable diseases (NCDs) are responsible for the deaths of 41 million people each year, which constitutes 74\% of all deaths globally\footnote{https://www.who.int/news-room/fact-sheets/detail/noncommunicable-diseases}. Cardiovascular diseases lead the list of NCD-related deaths, claiming 17.9 million lives annually. Therefore, the early detection, screening, and monitoring of NCDs are crucial for reducing their impact. Within this context, early and accurate detection of heart diseases is crucial for effective treatment. 

Traditionally, trained cardiologists and healthcare professionals manually examine electrocardiogram (ECG) readings. They interpret the waveforms to identify anomalies such as arrhythmias, ischemia, and other cardiac conditions. This process relies heavily on the expertise of medical professionals and can be time consuming, especially with large volumes of data. However, the trend is increasingly towards automation, leveraging machine learning (ML), deep learning (DL), and smart sensor technologies to improve the efficiency, accuracy, and real-time capabilities of heart disease monitoring. Automated systems can identify anomalies in ECG signals that might be indicative of heart conditions such as arrhythmias, ischemia, or myocardial infarction. This type of systems can be trained on large datasets to recognize patterns indicative of normal and abnormal heart activity \cite{anbalagan_analysis_2023,nezamabadi_unsupervised_2023}. In fact, by predicting these anomalies, it becomes possible to intervene sooner, potentially improving patient outcomes and saving lives. Moreover, these smart systems can even be employed in a hybrid fashion by providing decision support to clinicians \cite{jiang2024anomaly}. Thus, combining automated detection with manual review can enhance diagnostic accuracy and reduce the likelihood of false positives or negatives. 

When dealing with the recognition of ECG anomalies, current state-of-the-art models often demonstrate impressive accuracy and performance within controlled datasets \cite{goldberger2000physiobank, moody2001impact,tan2019icentia11k}. However, they often struggle to generalize effectively to new or unseen data sources \cite{liu2023self}. This gap highlights a significant limitation in their practical utility, as real-world deployment requires robust baseline models capable of: \textit{a)} adapting to diverse and dynamic environments and, \textit{b)} carrying out their training pipeline without strictly requiring a labeled dataset. Addressing this gap requires developing new domain adaptation techniques that can accommodate and handle variations in ECG signals, as well as pushing toward self-supervise methodologies that can address the lack of labeled data in real-life scenarios \cite{lai2023practical}. In addition, by integrating these techniques and methods into forecast pipelines, we can provide more reliable and accurate early detection of cardiac anomalies in a wider range of clinical and everyday monitoring scenarios, ultimately leading to better patient outcomes and greater applicability of these advanced diagnostic tools. 

This work responds to the need to improve the versatility of automated heart anomaly detection from raw, unlabeled ECG data, in order to enhance the robustness of detecting ECG beat and rhythm anomalies across various domains. To this end, we propose a forecasting DL architecture, which uses a novel morphologically inspired loss function to predict normal ECG signals. Moreover, we employ such a baseline model to further perform domain adaptation to other datasets. Finally, the forecasted ECG signals are compared with the actual sensor signals to detect anomalies. The latter is done by selecting a threshold distance that maximizes the separation between normal and anomalous samples.

In summary, our main contributions can be listed as follows:
 \begin{itemize}
     \item New DL model for ECG forecasting, which is trained in a self-supervised fashion with a novel morphological-inspired loss.
     \item Self-supervised domain adaptation process linked to a new distance metric to detect anomalies on ECG.
     \item Thorough empirical study to validate the performance of the proposed approach and an understandability study to unravel the knowledge acquired by the model.
     \item Accurate prediction of whether a sample is normal or contains an anomaly, despite our model being self-supervised and having never seen anomalous samples.
 \end{itemize}

Comparing the results of both experiments, we can observe that the domain-adapted model increases the performance a $12\%$, showing the high impact on the accuracy of adapting the model to the new domain. It is important to point out that despite our model being self-supervised and having never seen anomalous samples; it is able to accurately predict whether the sample is normal or contains an anomaly with a high degree of accuracy. Moreover, despite using 25 random combinations selected from the dataset, the standard deviation remains small, especially for the domain-adapted model, confirming the robustness of our approach.   

The paper is structured as follows: Section \ref{related} presents an overview of other systems in the literature that target ECG prediction and anomaly detection. Section \ref{sec:proposal} details the architecture of the system, focusing on every step involved in the pipeline. In Section \ref{methodstools}, we present the datasets and implementation details. Then, Section \ref{results} validates the proposed system by showing the experimental results and performing different analyses, including forecasting robustness, heart anomaly detection performance, temporal length input dependency, and model explainability. Finally, Sections \ref{discussion} and \ref{conclusions} discuss the main points of this work and offer our conclusions, respectively.

\section{Related Work}
\label{related}

Supervised heart anomaly detection approaches have relied heavily on labeled datasets to train models for heartbeat classification. These methods, while effective, face limitations due to the labor intensive process of obtaining labeled data and the variability of ECG signals between different patients and conditions \cite{anbalagan_analysis_2023}. Moreover, the lack of labeled data in real-life scenarios hinders the deployment of such supervised techniques. To address these challenges, recent research has increasingly focused on self-supervised and unsupervised DL for ECG analysis \cite{li_contrastive_2021,reviewSSLbiosignals,nezamabadi_unsupervised_2023}. Self-supervised learning, in particular, has gained attraction due to its ability to deal with unlabeled data and its effectiveness in leveraging different datasets to learn robust representations that improve downstream task performance. In general, this approach involves creating surrogate or pretext tasks that do not require manual annotations. Specifically, pretext methodologies can be divided into contrastive, predictive, and generative categories. In the context of ECG, different works have employed self-supervised learning frameworks for anomaly detection using such pretext task techniques \cite{reviewSSLbiosignals}. 

First, for the contrastive approach, although most of the recent works are targeting ECG-based arrhythmia recognition \cite{kiyasseh2021clocs,wei2022contrastive,lan2022intra,phan2022multimodality,jin2024self}, there are also some studies applying such methodology and derivatives for general cardiac anomaly detection. For instance, in \cite{zheng2022task}, a task-oriented self-supervised contrastive learning approach, initially devised for anomaly detection in electroencephalography, was proposed to tackle ECG anomaly detection. This method uses a 3-class convolutional neural network trained on artificially generated abnormal ECG data, simulating anomalies by varying waveform amplitude and frequency among individuals. This innovative approach was validated over an open dataset and reported a 75\% F1-score for a binary ECG anomaly detection task. However, depending on simulated anomalies brings up questions about the method's capacity to effectively represent the intricate and varied nature of real-world ECG irregularities, highlighting the need for a more suitable self-supervised learning approach. In a similar contrastive vein to the previous work, in \cite{mehari2022self}, the authors adapted state-of-the-art self-supervised methods, precisely instance discrimination and latent forecasting, to train models on 12-lead ECG data without heavily relying on labeled datasets. Utilizing public datasets, they demonstrated that pretrained models achieve near-supervised performance. Although this study establishes a compelling case for self-supervised techniques in biosignal representation learning and delves into some of the advantages of fine-tuning a self-supervised pre-trained model in a contrastive context, they are still relying on augmentations to generate semantically equivalent views of the original ECG data. The latter can sometimes introduce synthetic artifacts that do not fully capture the complexity and variability of real-world ECG signals. Secondly, the self-supervised predictive learning pipeline can be spotted in different works dealing with cardiac anomaly classification \cite{9630616,lee2021self,ZHANG2023104194,alamr2023unsupervised,jiang2024anomaly}. From these, \cite{jiang2024anomaly} stands out as a recent research effort to propose a self-supervised predictive learning system for general ECG anomaly detection. Specifically, the authors employed a trend-assisted method that is fed from a random masked multi-scale restoration to obtain an ECG trend feature. This is done following both a global and a local ECG analysis to end up with global-local ECG pairs features. Note that for the local restoration they perform heartbeat segmentation. Finally, they employed the ECG reports as attributes or labels following a self-supervised predictive scheme to guide the model training using the features extracted. Their results using open datasets reported an F1-score of 88.33\%, outperforming the state-of-the-art. However, they rely on the same dataset for both training and testing, and they do not give details on how the partition between both sets was done 
. Additionally, the model's complexity and the requirement for detailed patient-specific information may limit its applicability in settings where such comprehensive data is not readily available. A similar approach is presented in \cite{alamr2023unsupervised} where the authors proposed a novel model using transformer-based architecture for detecting anomalies in ECG signals. The main objective is to utilize the transformer's capability to capture long-distance dependencies in time series data, improving anomaly detection performance. The model, comprising an embedding layer and a transformer encoder, was tested on different datasets, achieving an 89.5\% accuracy and a 92.3\% F1 score for the MIT-BIH dataset. However, the need for substantial computational resources and careful configuration of the transformer's model parameters make such a system challenging to implement in practice. Finally, different self-supervised generative learning systems are also found pushing and going towards a fully unsupervised methodology \cite{dutta2021med,kite2024unlocking} but require a final supervised approach to detect anomalies. 
Note that, unlike the contrastive and the predictive approaches, generative pipelines learn to model the entire data distribution. This enables them to capture a wide variety of features and patterns in the data, leading to richer and more diverse representations. Thus, these systems can provide a better holistic understanding of the data, which can be useful for downstream tasks.

Within this context, self-supervised generative learning offers the adequate tools to train models on large amounts of unlabeled data, leading to systems that are then fine-tuned with a smaller amount of data, enhancing their generalization capabilities. The latter fact is key when facing real-life scenarios where there is a scarcity of labels. In fact, more efforts are needed to generate pipelines that are applicable to real-life settings. For example, both the authors in \cite{CARRERA2019482} and in \cite{zhang2023} presented systems that aim to provide domain adaptation features. The former does it in a user-specific manner, while the latter employs memory modules embedded into the proposed model that are dynamically updated to reflect changes in the data distribution.

As a result of the limitations discussed previously found in the literature in the context of self-supervised anomaly detection of the ECG, this work presents a self-supervised generative learning framework. Specifically, this proposal emerges as a particularly innovative solution by bringing a forecasting pretext upstream task together with a domain adaption process using a completely different dataset, to later being able to perform general and fully unsupervised cardiac anomaly detection.

\begin{figure*}[tbh]
\centering
\includegraphics{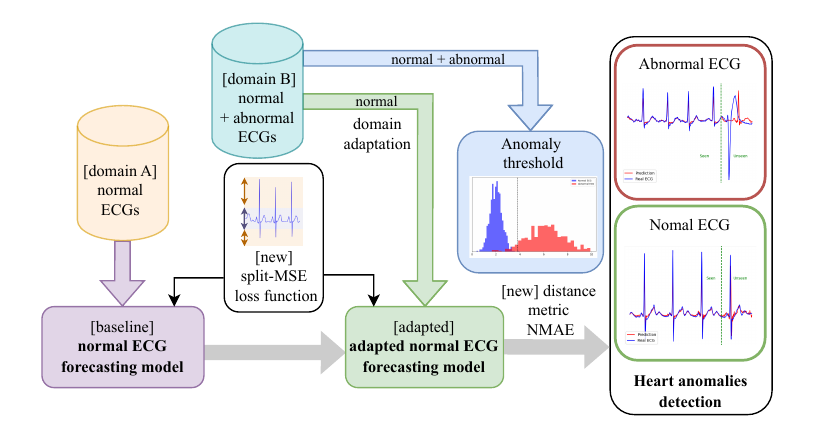}
\caption{\textbf{Proposed System.} General functionality of our proposed system for ECG forecasting with domain adaptation and heart anomaly detection.}
\label{fig:system}
\end{figure*}

\section{Methods} \label{sec:proposal}
A sketch of our proposed system is shown in Figure~\ref{fig:system}. We propose a forecasting model (purple box) to predict the future behavior of a normal (no containing anomalies) ECG signal. Training of our model is performed following a self-supervised approach that only requires a simple medical annotation of the input signals identifying normal ECG samples.    
As most of the time, people have normal ECG signals and their labeling by cardiologists is easier than classifying anomalies, there is much more data available, and this scheme allows us to train larger models with more generalization capabilities that can be deployed in different domains with good performance. In addition, we also propose a domain adaptation process (green box) to deal with new domains containing signals recorded with different leads, sensors, or sampling frequencies, that could potentially penalize the performance of our initial approach. The adaptation is also carried out through self-supervised training that employs normal ECG signals from the new domain. Finally, anomalies in the ECG signal are detected by comparing the forecasted values with real data (blue box).

\subsection{Input data} \label{sec:input}

\begin{figure}[tbh]
\centering
\includegraphics[width=0.6\columnwidth]{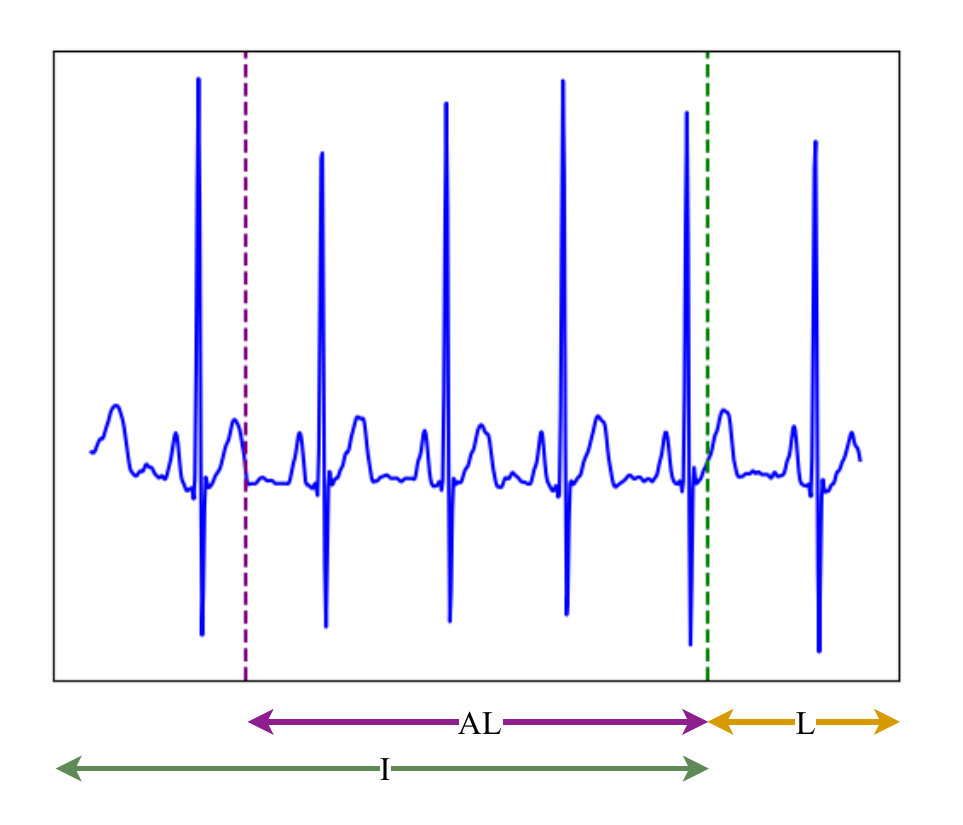}
\caption{\textbf{ECG signal slicing.} ECG slices for training the forecasting model: $I$ represents the input data, $L$ is the label data and $AL$ shows the auxiliary label data.}
\label{fig:input}
\end{figure}

The proposed forecasting system uses raw ECG signals as inputs and labels after applying typical filters, i.e. median and bandpass filters, to remove noise from signals. As indicated in Figure \ref{fig:input}, an input sample ($I$) has a length of $W_{I}$ and contains the input data used by the model to forecast the future signal. $W_{L}$ is the length of the label ($L$) which contains the future signal used to compare with the forecasted one. Furthermore, our system uses an auxiliary label ($AL$) containing $W_{AL}$ seconds from the end of the input sample ($I$) to stabilize the training process. More details on the different lengths are given later.




\subsection{Model Proposal}
\label{sec:model}

We propose a novel deep learning model for ECG forecasting (see Figure \ref{fig:intro}) whose final goal is to forecast $W_{L}$ seconds of normal ECG signals using $W_{I}$ seconds of normal ECG signals as input. Our model is inspired by the well-known SlowFast~\cite{feichtenhofer2019slowfast} network for video recognition and the U-Net \cite{ronneberger2015u} network for image segmentation. In this way, our architecture consists of two different parts: the ECG encoder and the ECG decoder. In addition, we also propose a novel training loss function that takes into account the specific morphological characteristics of the ECG signal.

\subsubsection{Model Encoder}

The encoder, which is inspired by SlowFast, codifies the ECG signal into a higher and richer dimensionality. It consists in turns of two pathways: the \textit{Slow} path (central path in Figure \ref{fig:intro}), operating at low frequency, which helps to understand coarse-grained details and high-resolution temporal information, and the \textit{Fast} path (right path in Figure \ref{fig:intro}), operating at high frequency, which allows the model to understand fine-grained details and low-resolution temporal information in the signal. Those paths are implemented using ResNet blocks together with the commonly used residual connections to avoid backpropagation issues \cite{he2016deep}. Due to the limited amount of available ECG data compared to the huge image or video datasets, our model uses an adapted and reduced number of layers. 

Before being passed to the different paths of the SlowFast, the ECG signal is first processed through a 1D Max Pooling layer with strides of two and one for the \textit{Slow} and \textit{Fast} paths, respectively, in order to extract the slow and fast frequencies, followed by a 1D Convolutional layer. Subsequently, four residual blocks compose every branch (see the Slow path in pink and the Fast one in green in Figure \ref{fig:intro}), and each block is composed of three ResNet bottlenecks. In order to train using our one-dimensional ECG signals, we have transformed the original 2D/3D convolutions into 1D convolutions; in this way, in our model, a ResNet bottleneck is formed by a succession of 1D convolutions and 1D batch-normalizations followed by a ReLU activation function. Moreover, the first bottleneck of each ResNet block has at the end a downsampling block, formed by one 1D convolution and one 1D batch-normalization, to reduce the temporal information. The output of the encoder is the concatenation of the outputs of both paths (see the pink and green combination in Figure \ref{fig:intro}), that is, the output of the slow path and the output of the fast path, with the aim of taking advantage of the final information from both.

\subsubsection{Model Decoder}

\begin{figure}[tbh!]
\centering
\includegraphics[width=0.87\columnwidth]{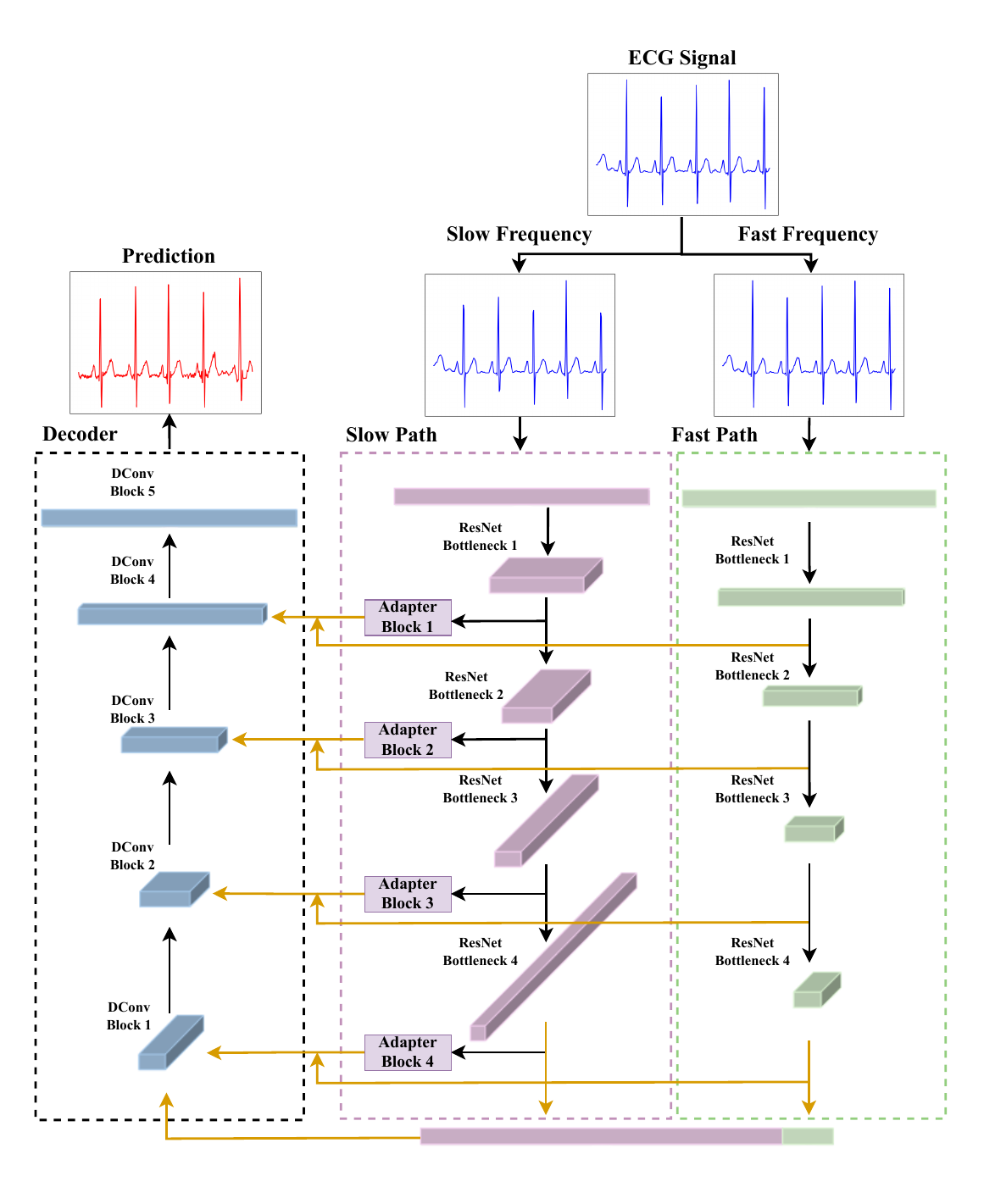}
\caption{\textbf{Proposed Model.} The proposed model for ECG forecasting that combines both SlowFast and U-Net networks is shown. The U-Net based decoder is on the left side (blue). The Slow path can be observed in the middle (color pink), while the Fast path is on the right side (green). Note that orange arrows indicate concatenation operation and black arrows show connections between layers. More details in the main text.}
\label{fig:intro}
\end{figure}

The decoder, which is inspired by U-Net, uses the encoder information to forecast the future ECG signal. It is composed of a succession of deconvolution blocks (left path in Figure \ref{fig:intro}). Following the design pattern of U-Net, the information from each residual block of the encoder is combined with the information from the deconvolution blocks (orange arrows in Figure \ref{fig:intro}) to take advantage of the information learned by the encoder. In this way, more robust signals are obtained at the end of the model. Note that in order to apply these deconvolution blocks, the outputs from the Slow and Fast path blocks are adapted to the same size to combine both sets of activations (adapter blocks in Figure \ref{fig:intro}). This adaptation is carried out by applying a 1D convolution, 1D batch-normalization, a ReLU activation function, and an upsampling step. Meanwhile, each deconvolution block consists of a combination of 1D deconvolution, 1D convolutions, and 1D batch-normalizations. Moreover, with the aim of preventing overfitting, dropout layers are also introduced between 1D batch normalizations and 1D convolutions. Finally, we attach a dense layer to predict the future signal that, as explained in Section~\ref{sec:input}, is composed of $W_{AL}+W_{L}$ seconds. In any case, only the $W_{L}$ forecasted seconds are used in the following steps of our pipeline.

\subsubsection{Morphological Loss function}
\label{sec:loss}

\begin{figure}[tbh]
\centering
\includegraphics[width=0.85\columnwidth]{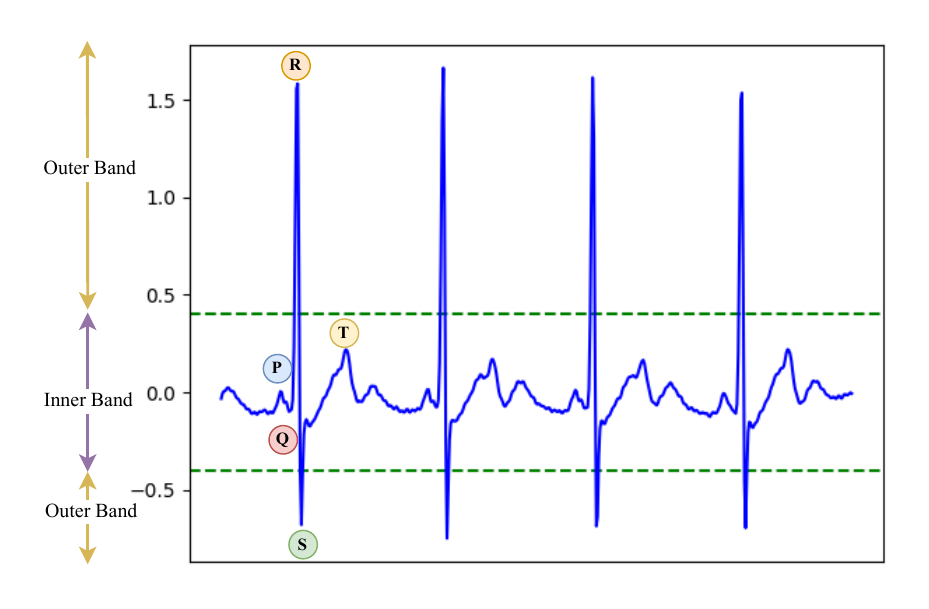}
\caption{\textbf{Split-MSE Loss Function.} MSE is calculated separately in two different ranges: outer bands, designed to capture the R-peaks, and inner band, which seeks to seize the segment between R-peaks. Note that P represents P-wave (atrial depolarization), QRS complex represents the depolarization of the ventricles and T represents T-wave (ventricular repolarization).}
\label{fig:loss}
\end{figure}

We present a novel loss function, named Split-MSE, specifically designed to deal with ECG signals. In this type of signal, we can observe: (i) the majority of the ECG values are close to 0, and only some peaks have higher values; (ii) the average of the signal is $\approx0$, leading to low loss values using traditional loss metrics such as mean squared error (MSE). Thus, we propose a variation of MSE as the loss function to train our model. Using Split-MSE, we divide the signal into three horizontal bands (see Figure \ref{fig:loss}). The outer bands are designed to capture the R peaks, whereas the inner band seeks to seize the segment between the R peaks (i.e., ST segment, PR interval, etc.). Mathematically, this loss function is calculated as described in Equation \ref{ec:mse}, where $\mathcal{L}_\text{MSE}^\text{inner}$ and $\mathcal{L}_\text{MSE}^\text{outer}$ are the MSE loss values calculated for the inner band and the outer band, respectively, and $w_{1}$ and $w_{2}$ are weights to balance values of the different ranges. Intuitively, increasing the value of $w_{1}$ causes the model to focus more on the inner band while increasing the value of $w_{2}$ causes the model to focus more on the outer band.


\begin{equation}
\mathcal{L}_\text{Split-MSE} = w_{1}\mathcal{L}_\text{MSE}^\text{inner} + w_{2}{L}_\text{MSE}^\text{outer}
\label{ec:mse}
\end{equation}

\subsection{Domain Adaptation}

\begin{figure}[tbh]
\centering
\includegraphics[width=1\columnwidth]{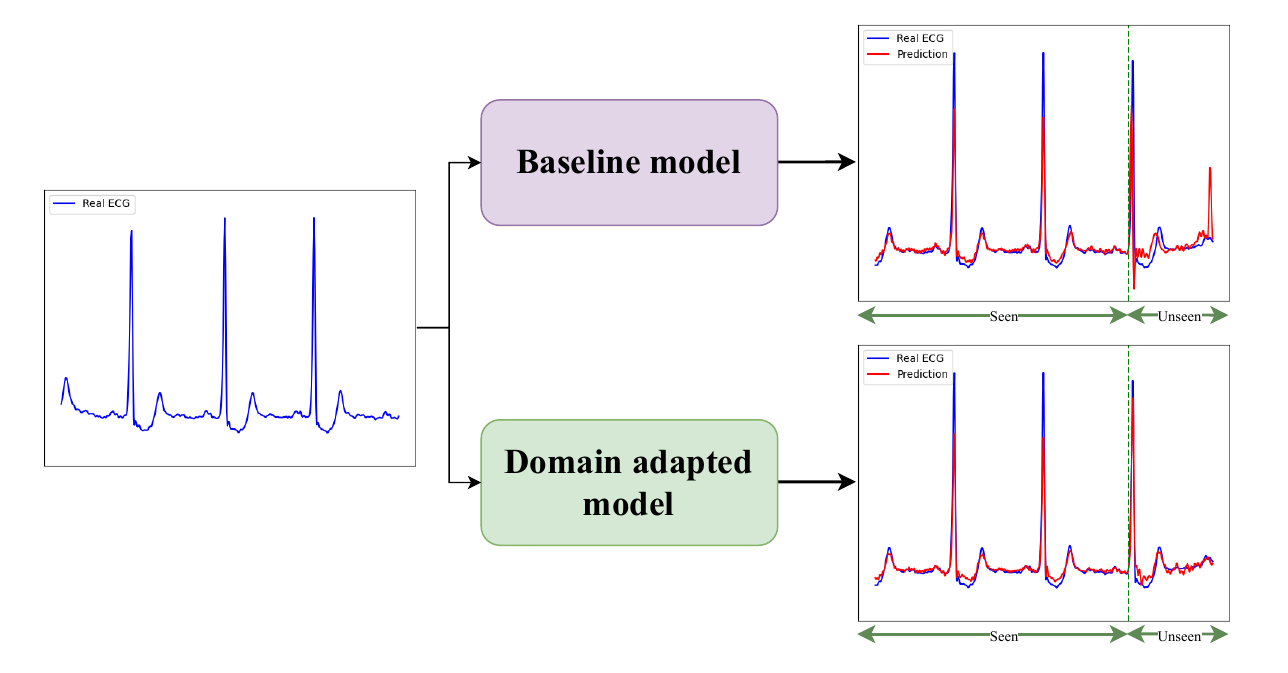}
\caption{\textbf{Domain Adaptation.} For a given sample of a new domain, the top row shows the prediction (unseen temporal segment) using the original model, while the bottom row represents the prediction employing the domain-adapted model. Huge differences can be appreciated in the unseen part of the signal. More details in the main text.}
\label{fig:domain}
\end{figure}

Once we have our forecasting model trained with normal ECG signals, it is crucial to adapt it to new contexts to accurately predict the behavior of signals recorded under different conditions caused by changes in leads, sensors, sampling frequency, and so on. In addition, subjects could have different lifestyles, stress levels, or even heart morphology that can affect the performance of our model. Therefore, it is necessary to fine-tune the previously trained model using samples from the new domain to deal with those changes and avoid a performance loss. This process is carried out following the same training process for the original forecasting model, that is, normal samples are used in a self-supervised way to predict the future signal. Thus, this domain-adapted model will be able to predict the future of normal input signals like the previously obtained model, but in a different domain. This improvement is illustrated with an example in Figure~\ref{fig:domain}. The upper row shows the prediction using the original model, whereas the lower row shows the prediction using the domain-adapted model. When examining the unseen (i.e., forecasted) segment of the signal, the original model erroneously produces an extra R-peak at the end of the predicted signal that is not present in the actual ECG. In addition, it generates a noisy signal at the beginning of the unseen segment. In contrast, the domain-adapted model avoids these issues, producing no extra R-peaks and yielding a cleaner signal than the baseline model.

\subsection{Anomaly detection} \label{sec:anomaly_detection}

Once the model is adapted to the target domain, the final goal is to detect heart anomalies from the input data. Since our model is trained to forecast normal ECG signals, we can compare the predicted sample with the real one received from the sensor to compute a distance metric. Thus, to detect anomalies, we simply need to establish a threshold to automatically classify a new sample as abnormal or normal. Notice that in an anomalous sample, the length subsequence $W_I$ has a normal heart behavior, but the length subsequence $W_L$ is anomalous.


We have designed a new metric, called NMAE, to evaluate the distance between real samples and predicted values. It combines the mean absolute error (MAE) between the raw label and the prediction, together with the mean absolute error of the label and the prediction, both normalized between 0 and 1, which is used as a weighting term. Mathematically, this distance metric is computed as described in Equation \ref{ec:mae}, where MAE is the standard mean absolute error, $W_L$ is the length of the signal, $L$ is the ECG signal used as label, $\widehat{L}$ is the predicted ECG signal, $L_{i}$ and $\hat{L}_{i}$ are the $i-th$ elements of the label signal and predicted signal, respectively.
\begin{center}
\footnotesize
\begin{align}
\text{NMAE} = \text{MAE} \cdot \frac{1}{W_{L}} \sum_{i=1}^{W_{L}} \left|\frac{L_i - \min(L)}{\max(L) - \min(L)} - \frac{\hat{L}_i - \min(\hat{L})}{\max(\hat{L}) - \min(\hat{L})}\right|
\label{ec:mae}
                   \end{align}
\end{center}

To determine the optimal threshold, we employ a systematic search strategy. Specifically, we select an unseen subset of abnormal and normal samples during training and calculate their corresponding NMAE distance with the predicted signal. These distances are expected to be lower for normal samples and higher for abnormal ones. Then, we iteratively evaluate candidate thresholds by calculating their classification accuracy across the two groups. 
The optimal threshold is the one that maximizes this classification accuracy. Mathematically, the global accuracy is calculated as described in Equation \ref{ec:acc}, where $D_N$ is the set of distances of normal samples, $D_A$ is the set of distances of abnormal samples, $T$ denotes the threshold, $\lvert D_N \rvert$ represents the total number of normal samples, $\lvert D_A \rvert$ represents the total number of abnormal samples, $\lvert \{ d_N \in D_N \mid d_N < T \} \rvert$ is the number of samples classified as normal and $\lvert \{ d_A \in D_A \mid d_A \geq T \} \rvert$ is the number of samples classified as abnormal.

\begin{align}
\text{Accuracy} = \frac{\frac{\lvert \{ d_N \in D_N \mid d_N < T \} \rvert}{\lvert D_N \rvert} + \frac{\lvert \{ d_A \in D_A \mid d_A \geq T \} \rvert}{\lvert D_A \rvert}}{2}
\label{ec:acc}
\end{align}


\section {Methodology and tools}
\label{methodstools}
In this section, we describe the ECG datasets that we use in our experiments, as well as the specific implementation particularities followed to train and test our proposed model.  

\subsection{Datasets}
To evaluate our approach, we employ two widely used datasets. Thus, our forecasting model is trained with the MIT-BIH Normal Sinus Rhythm database~\cite{goldberger2000physiobank}, which only includes normal samples. In addition, the domain adaptation approach is performed using the MIT-BIH arrhythmia database~\cite{moody2001impact}, which contains both normal samples and samples with heart anomalies. Next, a brief description of both databases is given.

\textit{MIT-BIH Normal Sinus Rhythm (NSR) database~\cite{goldberger2000physiobank}.} The subjects included in this database were found to have no significant arrhythmias. This dataset comprises 18 long-term ECG recordings, each lasting approximately 24 hours and sampled at 128 Hz, of subjects referred to the Arrhythmia Laboratory at Boston's Beth Israel Hospital (now the Beth Israel Deaconess Medical Centre). It includes 5 men aged 26 to 45 and 13 women aged 20 to 50.

\textit{MIT-BIH Arrhythmia database~\cite{moody2001impact}}. The subjects included in this database were found to have medical annotations indicating heart anomalies or normal ECG. This dataset contains 48 half-hour excerpts of two-channel ambulatory ECG recordings sampled at 360 Hz. These ECGs were obtained from 47 subjects studied by the Arrhythmia Laboratory at Boston's Beth Israel Hospital between 1975 and 1979. The subjects included 25 men aged 32 to 89 years and 22 women aged 23 to 89 years. At least, two cardiologists annotated each record independently, identifying not only the type of rhythm but also the type of beat. The ECG lead varied among subjects due to surgical dressings and anatomical variations that do not permit the use of the same electrode placement in all cases. In most records, one channel is a modified limb lead II (MLII), obtained by placing the electrodes on the chest, and the other channel is usually V1 (sometimes V2, V4, or V5, depending on the subject).

\subsection{Implementation Details}

Next, we will introduce the implementation details for training and testing our models. First of all, we will explain the preprocessing applied to our datasets. Then, we will describe the models training process. 

\subsubsection{Input data}
\label{sec:data}

We split all ECG recordings into windows of a length of four seconds, that is, $W_{I}=4$. The label has a length of 1 second, that is $W_{L}=1$, and the auxiliary label ($AL$) overlaps 3 seconds, that is $W_{AL}=3$, with the input signal. Note that the 1 second labels have been chosen because the normal heart rate of a clinically healthy person is established between 60 and 100 beats per minute \cite{ostchega2011resting,zhang2016resting}. Therefore, at least one second is usually necessary to have one beat and detect anomalies in the heart accurately. Finally, a specific preprocessing has been applied to each dataset, as follows: 

\textit{MIT-BIH NSR Preprocessing Details.}
We applied both median and bandpass filters to the signals, as previously discussed in Section \ref{sec:input}. Specifically, the median filter has a kernel size of 3, and the frequency range of the bandpass filter is between 0.5 Hz and 30 Hz. Furthermore, given that this dataset contains noisy data, we utilize ECG\_QC\footnote{\url{https://aura-healthcare.github.io/ecg_qc/}} library to classify ECG signals as good or bad quality using machine learning techniques. Specifically, we use a decision tree classifier to evaluate the signal quality every second and calculate the mean value for the entire input sample with the aim of removing excessively noisy samples. Taking into account this signal quality score, we retain only those samples with a score of 0.5 or higher, that is, at least 50\% of the signal conforms to the structure of a normal ECG. Finally, we normalize all samples to a common range of values to improve model training and remove outliers. Thus, based on the histogram of the maximum and minimum values of the signals in the dataset, we observed that approximately 80\% of the signals have a maximum value of 2.5 or lower and a minimum value of -0.75 or higher. Taking into account these maximum and minimum values, we remove all signals whose maximum value is 50\% higher than the established maximum or 50\% lower than the established minimum, i.e., higher than 3.75 or lower than -1.125. Ultimately, all signals are clamped within the range [-0.75, 2.5] and normalized between 2.5 and -0.75, separately for negative and positive values. Note that all these values were selected following a cross-validation empirical setup.

\textit{MIT-BIH Arrhythmia Preprocessing Details.} 

Provided that this dataset contains medical annotations for the type of rhythm and type of beat, and the ECGs are recorded using a variety of leads, we have selected a subset of it based on three criteria: (i) to unify the type of signal used to train our forecasting model, we selected only those subjects whose ECGs were recorded using a modified limb lead II (MLII), obtained by placing the electrodes on the chest; (ii) since our approach requires a temporal window of normal signal, we removed those subjects whose all samples contain abnormal beats or abnormal rhythms; (iii) due to the input requirements of our model, we only keep those anomalies whose previous four seconds were normal. These three criteria resulted in the selection of 35 subjects\footnote{100, 101, 103, 105, 106, 107, 108, 112, 113, 114, 115, 116, 117, 119, 121, 122, 123, 200, 201, 202, 205, 208, 209, 212, 213, 215, 217, 219, 220, 222, 223, 228, 230, 233 and 234} and a total of 26628 samples.

In addition, resampling was performed to match the frequency with that of the MIT-BIH NSR dataset, reducing it from 360 Hz to 128 Hz. The median and band-pass filters were applied with the same configuration as for the previous dataset. Given that this database comprises normal and anomalous signals, unlike the MIT-BIH NSR dataset, signal quality assessment and normalization procedures were not applied in order to maintain maximum fidelity with respect to the original anomalous signals.

\subsubsection{Training details}
As previously mentioned, the forecasting model is trained using the MIT-BIH NSR dataset to develop a generic model capable of predicting the future of NSR signals. We divided the dataset into a training set and a test set following a Leave-One-Subject-Out (LOSO) strategy. This way, we employ the ECG of one random patient for testing the model and the rest of the subjects for training. Note that this dataset is used just to train the baseline forecasting model, whose aim is the prediction of the future behavior of NSR signals. As anomaly detection will be performed on another dataset after having applied a domain adaption approach, we save computational time and resources by performing the LOSO strategy only once with a random subject (to enable the replicability of our experiments, we want to clarify that we have used Patient number 5) to check that the model is working properly. For training this model, we performed 200 epochs with a learning rate starting at $1\times 10^{-4}$ and ending at $1\times 10^{-7}$ using a cosine function over the 200 epochs. We employ AdamW optimizer with batches of 256 samples and weight decay of $1\times 10^{-5}$. During training a dropout of $20\%$ has been established with the objective of avoiding overfitting. Regarding the Split-MSE loss function (Equation \ref{ec:mse}), the band limits are set at $\pm0.4$, hence the inner band comprises the range $[-0.4, 0.4]$ while the outer band is defined as $(-\infty, -0.4) \cup (0.4, \infty)$, for $\mathcal{L}_\text{MSE}^\text{inner}$ and $\mathcal{L}_\text{MSE}^\text{outer}$, respectively, with $w_{1}= 5$ and $w_{2}=1$. Note that all these hyperparameters have been selected following a cross-validation empirical setup. Our proposed model comprises a total of approximately 4.9 million parameters that require 9.78 MB of memory allocation during the forward pass. Additionally, the memory usage of the parameters is 19.56 MB, which in turn leads to a total memory consumption of 29.34 MB. We conducted our experiments on a server composed of an Intel Xeon Silver 4314 processor with 64 cores, running at 2.4 GHz, 500 GB of RAM, and four GPUs, GeForce RTX 3090 with 24 GB. All experiments were performed using Python 3.10 with PyTorch 2.1.0 running on CUDA 11.8.

\subsubsection{Domain Adaptation details}
\label{sec:domain_adapt_details}
The domain adaptation is carried out using the MIT-BIH Arrhythmia dataset to adapt the forecasting model to a different context (i.e., different sensors, environmental conditions, heart morphology, etc.). Notice that this dataset has both normal and abnormal samples, but only normal samples are used for domain adaptation. In practice, we have randomly shuffled normal samples and divided them according to a ratio of $80\%$ training samples and $20\%$ test samples. To adapt the model to the new context, we fine-tuned 250 epochs with a learning rate starting at $1\times 10^{-4}$ and ending at $1\times 10^{-8}$ using a cosine function over the 250 epochs. We employ AdamW optimizer with batches of 64 samples and weight decay of $1\times 10^{-5}$. During training a dropout of $20\%$ has been established with the objective of avoiding overfitting. Regarding the Split-MSE loss function (Equation \ref{ec:mse}), the band limits are set at $\pm0.3$, hence the inner band comprises the range $[-0.3, 0.3]$ while the outer band is defined as $(-\infty, -0.3) \cup (0.3, \infty)$, for $\mathcal{L}_\text{MSE}^\text{inner}$ and $\mathcal{L}_\text{MSE}^\text{outer}$, respectively, with $w_{1}= 1$ and $w_{2}=1$. Again, all these hyperparameters have been selected following a cross-validation empirical setup. Note that the training hyperparameters differ between the forecasting model and the domain adaptation model due to a fine-tuning process is applied to the first model, e.g. a lower learning rate is required. Moreover, the datasets are different, necessitating adjustments such as different batch size and modified band limits.

\subsubsection {Anomaly detection details}
\label{sec:anomaly_detect_details}
For anomaly detection, we use all abnormal samples from the MIT-BIH arrhythmia dataset and all normal samples included in the test set of the domain adaptation approach, as explained in the previous section. In this way, the anomaly detection approach employs samples that have not been used to train the domain-adapted model. With these samples, we must build a training set composed of the $20\%$ of normal and abnormal samples available to calculate the anomaly detection threshold value as explained in Section \ref{sec:anomaly_detection}. We search for the threshold by evaluating percentiles of the anomalous distance values in steps of $1\times 10^{-4}$ within the range from 0 to 100. The optimal threshold is the one that maximizes the global accuracy. The remaining values compose the final test set that is used to calculate the accuracy of the anomaly detection task.

\section{Experimental results}
\label{results}

In this section, we present the experiments conducted to evaluate both the forecasting and detecting heart anomalies capabilities of our approach. We want to highlight that, despite the existence of other recent works \cite{alamr2023unsupervised,kite2024unlocking} that apply a self-supervised generative learning scheme, they did not specify their data partitioning methods and sampling strategy. Consequently, it is not possible to compare them directly with our proposal. However, we consider that this comparison is important, so that we have trained the transformer architecture described in \cite{alamr2023unsupervised} using our dataset. Therefore, our experiments focus on evaluating the performance of the different components proposed in Section~\ref{sec:proposal} and a comparison with the transformer model. Note that in contrast to previous work, we have specified how to exactly split the datasets and use the data to facilitate future comparisons with our approach. Moreover, the complete code of our proposal will be publicly available online\footnote{\url{https://github.com/PaulaRuizB/FADE-ECG}}. Focusing on the experiments, they are divided into six sections: forecasting model, domain-adapted model, heart anomaly detection, temporal length study, model understandability and ablation study.

\subsection{Forecasting Model}

\begin{figure}[tbh]
\centering
    \begin{subfigure}[b]{0.47\textwidth}
    \centering
    \includegraphics[width=\textwidth]{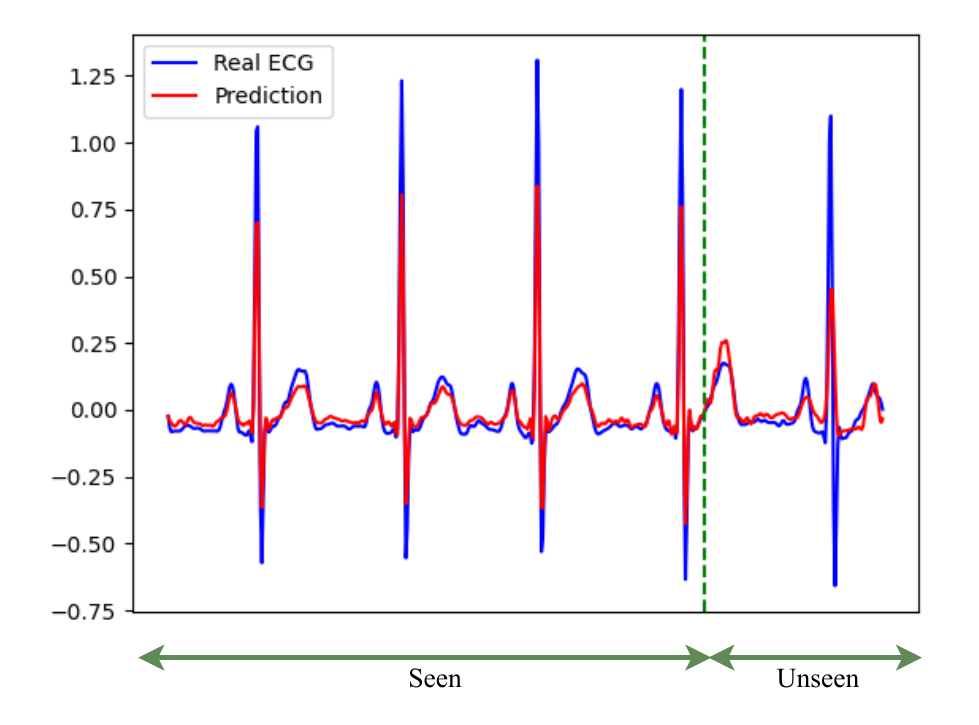}
    \caption{ECG forecasting using MSE\label{fig:mse}}
    \end{subfigure}
\quad
    \begin{subfigure}[b]{0.47\textwidth}
    \centering
    \includegraphics[width=\textwidth]{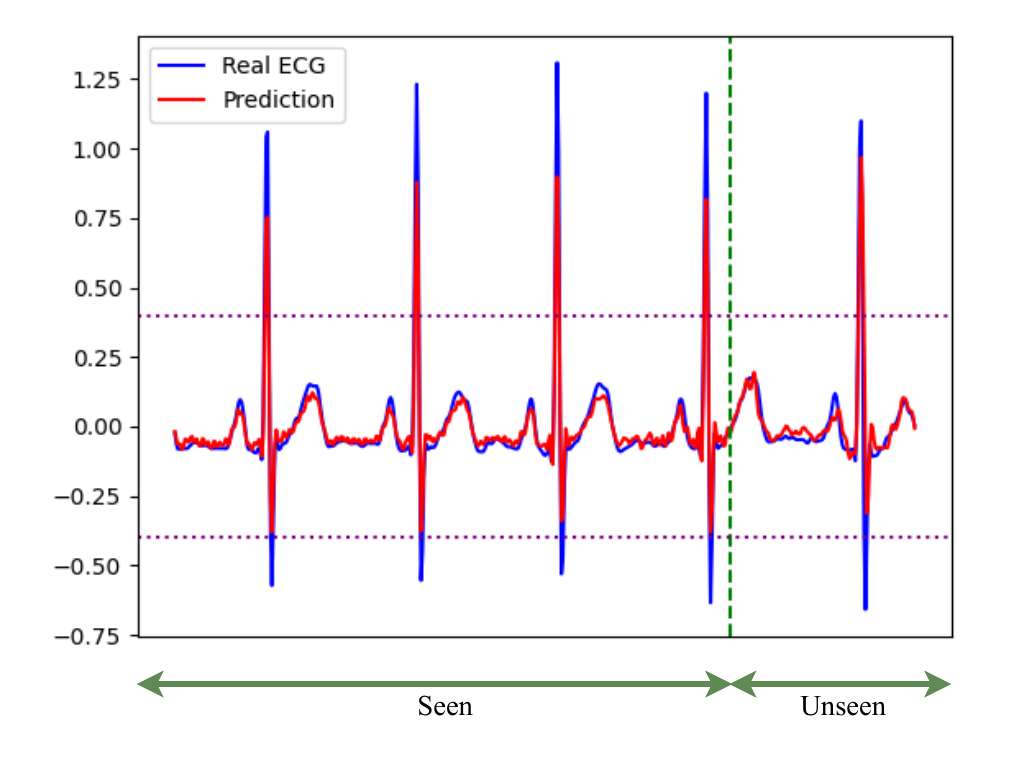}
    \caption{ECG forecasting using Split-MSE\label{fig:mse_in}}
    \end{subfigure}
\caption{\textbf{Example of an ECG forecasting comparison for the same input.} (a) ECG forecasting trained with MSE. (b) ECG forecasting trained with Split-MSE. Vertical dash-green lines separate the previously seen signal in the input and the new second unseen by the model. Purple dotted lines indicate the limits for inner and outer MSE calculation.}
\label{fig:mse_mse_in}
\end{figure}

In this subsection, we evaluate the prediction performance of the proposed forecasting model using a test set from the MIT-BIH NSR dataset, which comprises approximately 75000 signals from a randomly selected patient (in our case, patient number 5). To measure the performance of the model, we use our NMAE metric proposed in Section~\ref{sec:anomaly_detection}, taking into account that for this metric, the lower the value, the better. In this experiment, we have evaluated our proposed model with different numbers of blocks and different loss functions, but for brevity, we will focus on the best cases. Since the dataset is large enough to provide a wide variety of samples, we did not observe an accuracy drop when the number of blocks is increased; it just tends to stabilize. Thus, we selected the model composed of 4 blocks that provided a good trade-off between accuracy and the amount of parameters. Regarding the loss functions, we have evaluated the traditional MSE and our Split-MSE proposed in Section~\ref{sec:model}. Note that additional loss functions were tested, but their convergence was worse than expected, so we commit those values for brevity. Comparing the standard MSE loss with our Split-MSE, the first one obtained an NMAE value of 0.0136, while the second one achieved 0.011. Both values are close to zero due to the characteristics of the ECG signal, but Split-MSE obtains a lower distance value, which is better using our distance metric.  Figure~\ref{fig:mse_mse_in} illustrates the impact of using both losses in the forecasted model. We can observe that R-peaks were poorly predicted with MSE loss (Figure~\ref{fig:mse}) while our Split-MSE (Figure~\ref{fig:mse_in}) provides better predictions, especially in the second segment never seen by the model. From these values, we can conclude that the model trained with the Split-MSE loss function predicts signals that are more similar to the real signals than those predicted by the model trained with the MSE loss function.

To compare our approach with other state-of-the-art forecasting models, we have trained and tested the transformer architecture described in \cite{alamr2023unsupervised} under the same experimental setup to carry out a fair comparison. This publication is particularly relevant because they utilize the same MIT-BIH Arrhythmia dataset and report achieving an accuracy of $89.5\%$ with their proposed approach. For brevity, we report the results of the transformer model with two encoder blocks, 32 heads and a hidden size of 64, which is the configuration that achieves the best accuracy in the original publication. However, as they did not specify other details, such as token creation, we have trained this transformer using our dataset processing. Note that additional setups have been tested but we only report the best one. To compare the performance between our proposed forecasting model and the transformer model presented in \cite{alamr2023unsupervised} we obtain the NMAE values. Our model obtained an NMAE value of $0.011$ whereas the transformer obtained an NMAE value of $0.022$. From these values, we can conclude that our model predicts better future signals than the transformer model. This may be due to the fact that in the original paper, this model uses segmented parts of the signal as input, while we do not conduct any segmentation of the signal, providing a more realistic environment.

\subsection{Domain-adapted model}

The performance of the domain-adapted model can be measured using a qualitative approach by comparing the resulting signals generated by the model or employing a quantitative approach by calculating the accuracy obtained during anomaly detection. In this section, we follow the former approach, while in the next section we develop the latter.

\begin{figure*}[tbh] 
    \centering
    
    \begin{subfigure}[t]{0.24\textwidth} 
        \centering
    \includegraphics[width=\textwidth]{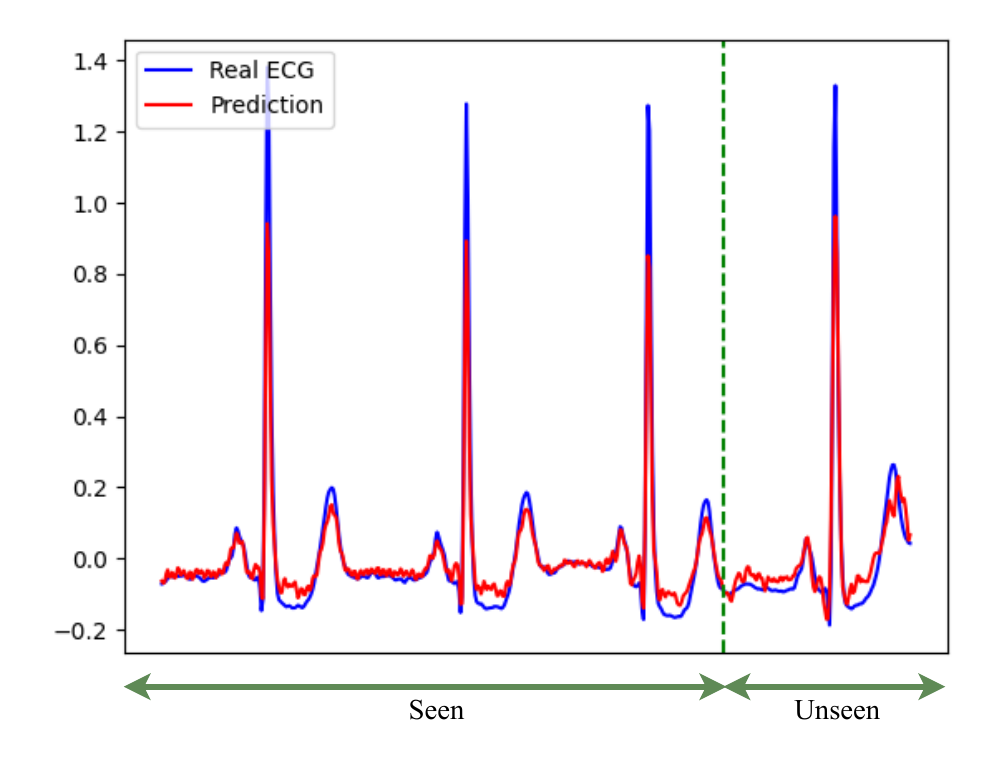} 
        \caption{Normal ECG Patient 101}
        \label{fig:subfigura1}
    \end{subfigure}
    \begin{subfigure}[t]{0.24\textwidth}
        \centering
    \includegraphics[width=\textwidth]{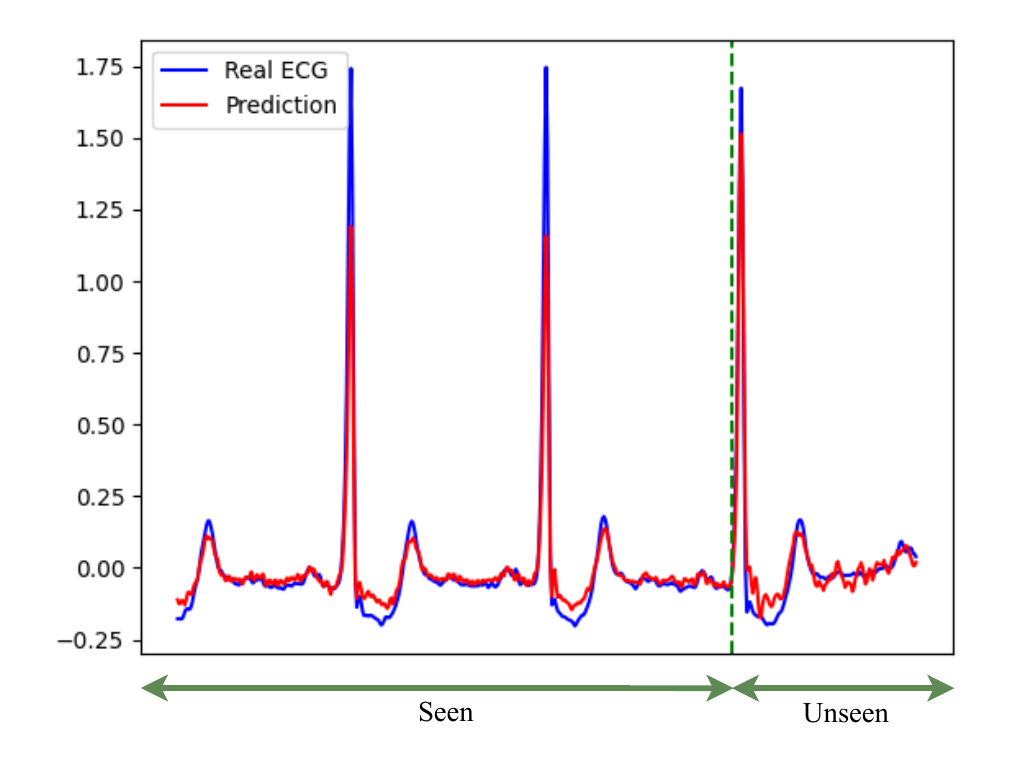}
        \caption{Normal ECG Patient 114}
        \label{fig:subfigura2}
    \end{subfigure}
    \begin{subfigure}[t]{0.24\textwidth}
        \centering
        \includegraphics[width=\textwidth]{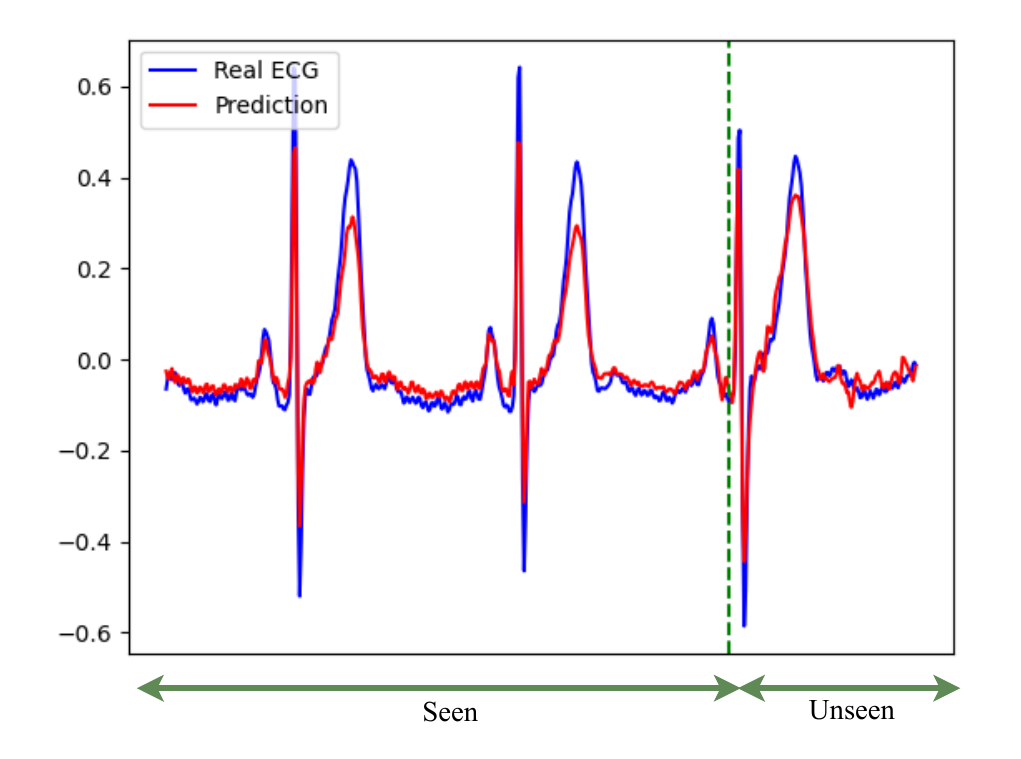}
        \caption{Normal ECG Patient 117}
        \label{fig:subfigura3}
    \end{subfigure}
    \begin{subfigure}[t]{0.24\textwidth}
        \centering
        \includegraphics[width=\textwidth]{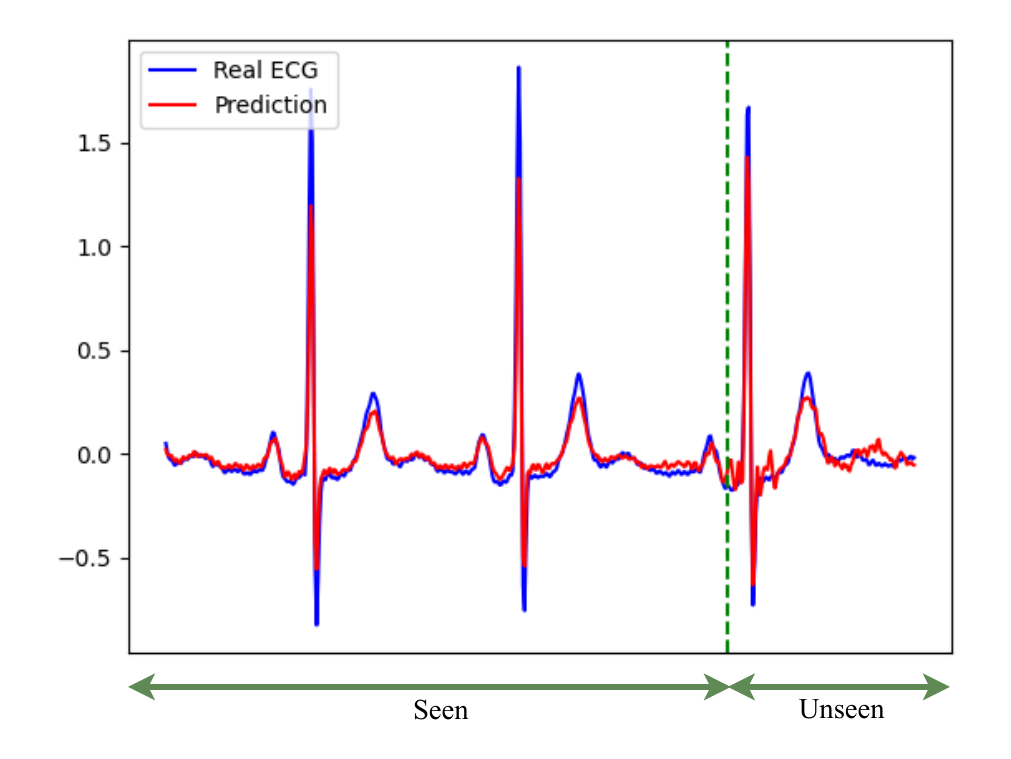}
        \caption{Normal ECG Patient 123}
        \label{fig:subfigura4}
    \end{subfigure}
    
    \vskip\baselineskip
    \begin{subfigure}[t]{0.24\textwidth} 
        \centering
        \includegraphics[width=\textwidth]{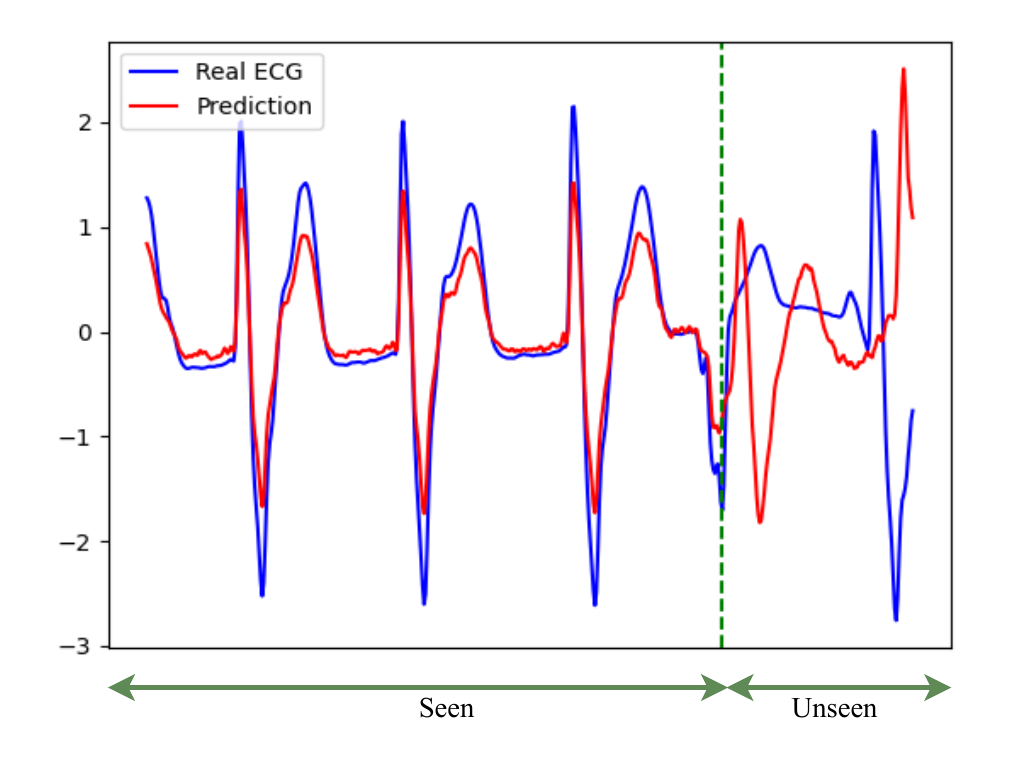} 
        \caption{Premature Ventricular Contraction Patient 107}
        \label{fig:subfigura5}
    \end{subfigure}
    \begin{subfigure}[t]{0.24\textwidth}
        \centering
        \includegraphics[width=\textwidth]{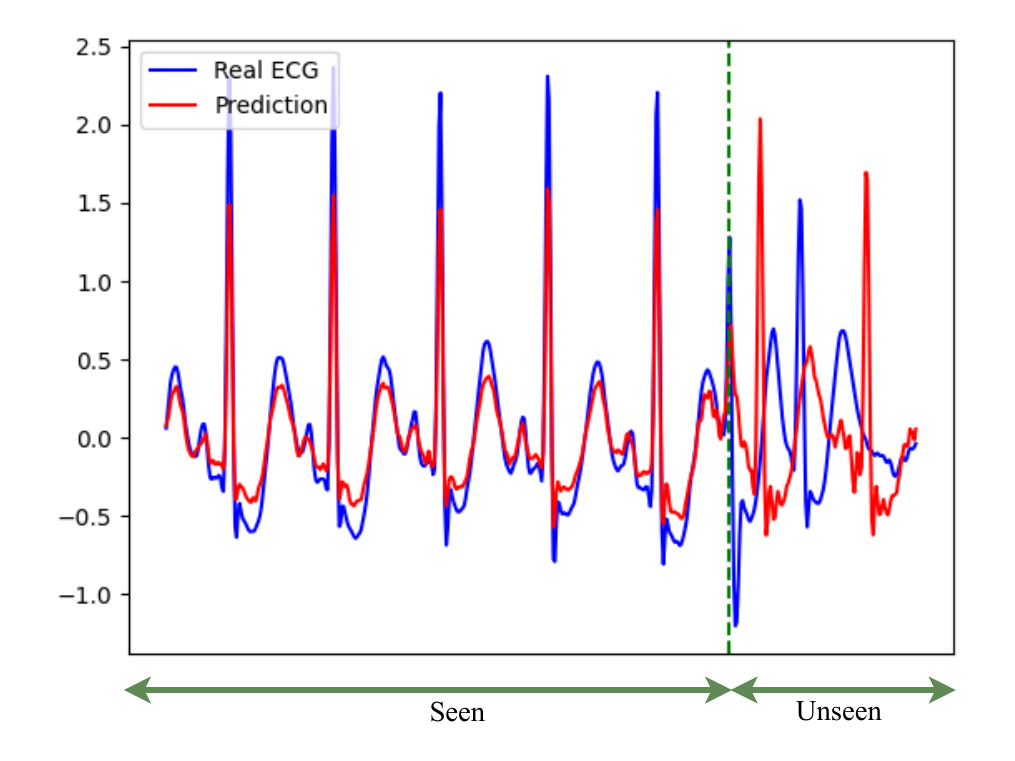}
        \caption{Aberrated Atrial Premature Beat Patient 213}
        \label{fig:subfigura6}
    \end{subfigure}
    \begin{subfigure}[t]{0.24\textwidth}
        \centering
        \includegraphics[width=\textwidth]{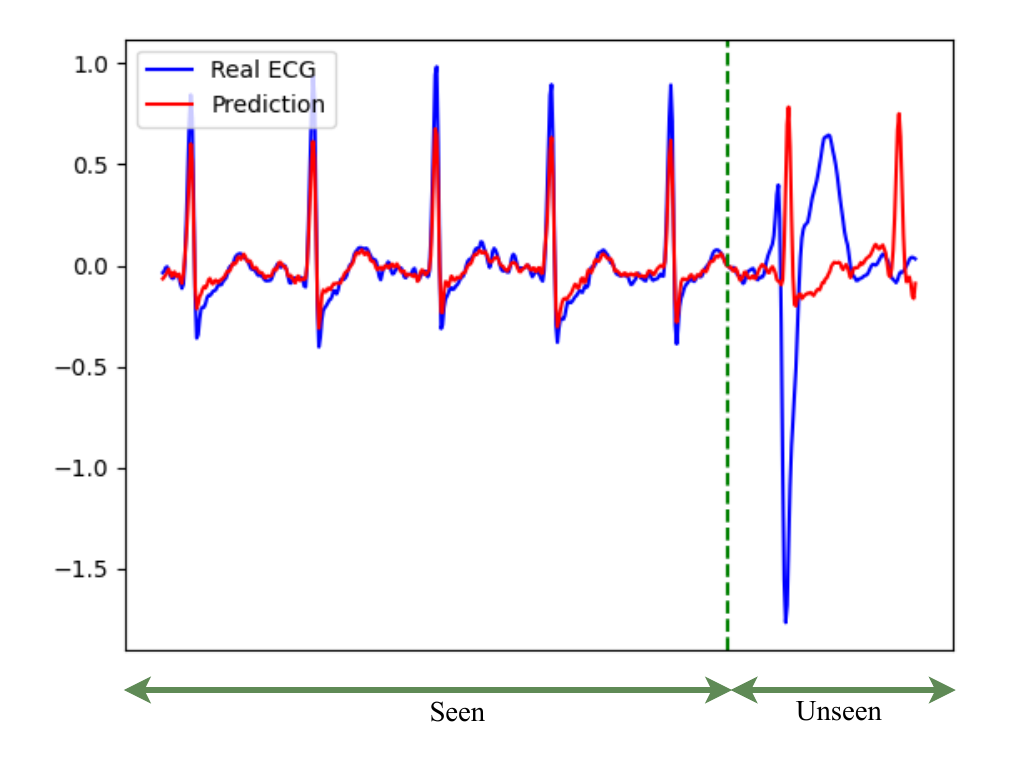}
        \caption{Ventricular Bigeminy Patient 200}
        \label{fig:subfigura7}
    \end{subfigure}
    \begin{subfigure}[t]{0.24\textwidth}
        \centering
        \includegraphics[width=\textwidth]{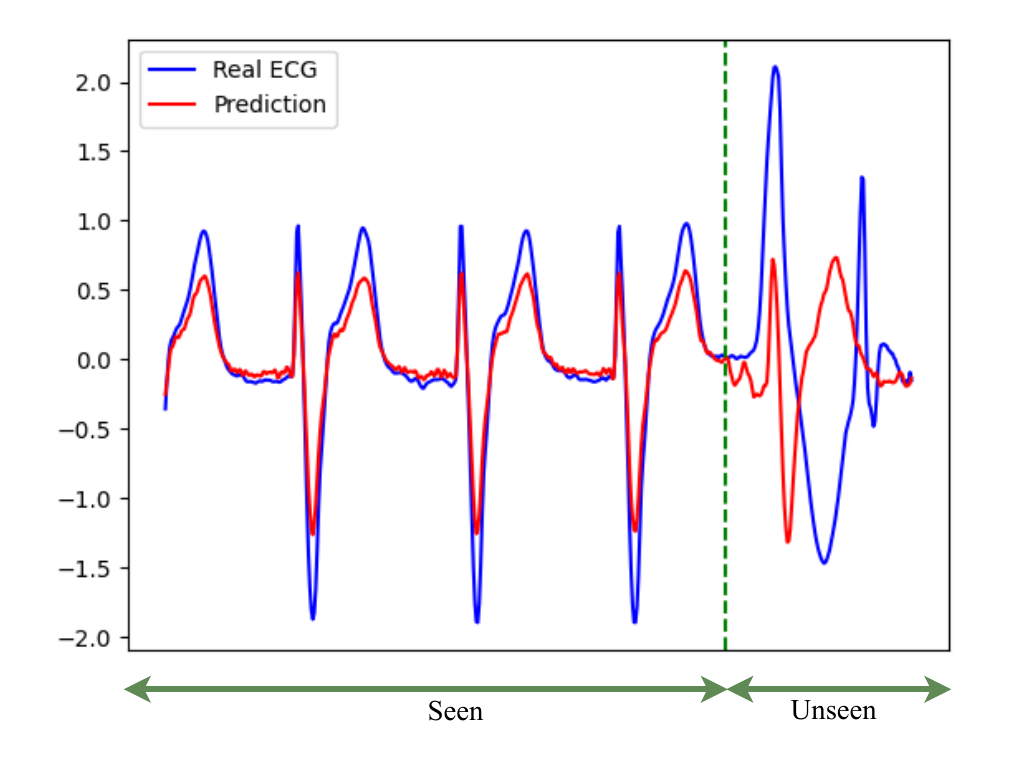}
        \caption{Ventricular Tachycardia Patient 217}
        \label{fig:subfigura8}
    \end{subfigure}
    
    \caption{\textbf{Examples of ECG forecasting for different patients.} The top row shows normal signals, while the bottom row shows anomalous signals. Note that blue represents the real ECG and red represents the predicted signal.}
    \label{fig:signals}
\end{figure*}

Thus, here we present different visual examples of predicted signals utilizing our domain-adapted ECG forecasting model. On the one hand, we forecast normal ECG signals, that can be observed in the top row of Figure \ref{fig:signals}. In Figures \ref{fig:subfigura1} to \ref{fig:subfigura4}, it can be seen that the last unseen second is considerably well predicted, even though the ECGs belong to different people. These signals are slightly different from one person to another, not only in form but also in amplitude. For example, the maximum value in the signal from Patient 117 (Figure \ref{fig:subfigura3}) is approximately 0.6, while that for Patient 114 (Figure \ref{fig:subfigura2}) is approximately 1.75. Thus, we can conclude that our model is capable of adapting to very different types of normal ECG signals.
On the other hand, we utilize ECG signals whose last second is an anomaly, as shown in the bottom row of Figure \ref{fig:signals}. In Figures \ref{fig:subfigura5} to \ref{fig:subfigura8}, it can be observed that our model predicted the unseen second as a normal signal (colored red), whereas the real ECG has an anomaly (colored blue). Therefore, the distance is high and the signals are classified as anomalous ECGs. In Figures \ref{fig:subfigura5} and \ref{fig:subfigura6}, the anomaly arises in the type of beat, while in Figures \ref{fig:subfigura7} and Figure \ref{fig:subfigura8}, it occurs in the type of rhythm. Therefore, our system is able to detect anomalies in the ECG regardless of the type of anomaly. 

\subsection{Heart Anomalies Detection}

Two types of experiments are conducted with respect to anomaly detection in the MIT-BIH Arrhythmia dataset. The first experiment uses the domain-adapted model as explained in Section \ref{sec:domain_adapt_details} and works with the training and test sets defined in Section \ref{sec:anomaly_detect_details}. We follow a five-fold cross-validation process to evaluate the consistency of the obtained accuracy values. In this way, we randomly split the available MIT-BIH arrhythmia normal samples for training and testing five times. After that, we must apply a second five-fold split over the test set for the threshold selection, including normal and abnormal samples, resulting in a total of 25 dataset combinations and 25 values of accuracy. The average accuracy achieved by the domain-adapted model is $84.65\pm0.56\%$, with average accuracies for anomalies and normal samples of $83.84\pm2.97\%$ and  $85.46\pm2.99\%$, respectively.

In the second experiment, we quantify the advantages of using a domain-adapted model. Thus, the non-adapted model is used to forecast the ECG signals following the same test setup used for the adapted model. We obtain an average accuracy of $72.43\pm0.45$, with average accuracies for anomalies and normal samples of $68.27\pm6.88$ and $76.60\pm5.99$, respectively.

Comparing the results of both experiments, we can observe that the domain-adapted model increases the performance a $12\%$, showing the high impact on the accuracy of adapting the model to the new domain. It is important to point out that despite our model being self-supervised and having never seen anomalous samples; it is able to accurately predict whether the sample is normal or contains an anomaly. Moreover, despite using 25 random combinations selected from the dataset, the standard deviation remains small, especially for the domain-adapted model, confirming the robustness of our approach.    

Finally, we want to highlight that we have tested several distance metrics and compared their performance with that of the model in terms of accuracy. Table \ref{tab:metrics} shows the accuracy of the best ten metrics we have evaluated using the domain-adapted model. Moreover, we have used the weighting term in these metrics as in our proposed metric (see Section \ref{sec:anomaly_detection}), but none of them has achieved higher accuracy than NMAE, as can be seen in Table \ref{tab:metrics}. Thus, NMAE is the distance metric used in all experiments.

\begin{table}[tbh]
\centering
\scriptsize
\begin{tabular}{lccc}
\hline
\textbf{Distance Metric}       & \textbf{Global (\%)} & \textbf{Anomalies (\%)} & \textbf{Normal (\%)} \\ 
\hline
 NMAE &   \textbf{85.45}              &       \textbf{83.95}           &     \textbf{86.95}         \\ 
MAE &     83.65            &       80.59             &     86.71         \\ 
SMAPE &      83.59           &        81.95            &      85.22        \\ 
 Bray-Curtis &    79.39             &     72.82               &    85.96          \\ 
LCE &       77.45          &       73.98                &       80.92       \\ 
SmoothL1 &       76.88         &       69.78             &    83.98          \\ 
MSE &       76.27           &       77.23             &     75.31         \\ 
Correlation &    72.36             &    79.33                &   65.40           \\ 
 RSE &     72.23            &        82.37            &     62.10         \\ 
 Chebyshev  &     66.12            &     81.53               &      50.70        \\ 
\hline
\end{tabular}
\caption{\textbf{Accuracy using different distance metrics.} Global accuracy, anomalies accuracy and normal signals accuracy for: normalized mean absolute error (NMAE), mean absolute error (MAE), symmetric mean absolute percentage error (SMAPE), Bray-Curtis, logarithm of the hyperbolic cosine of the prediction error (LCE), SmoothL1, mean squared error (MSE), correlation, relative squared error (RSE) and Chebyshev.}
\label{tab:metrics}
\end{table}

Returning to the comparison of our proposal with the transformer model presented in \cite{alamr2023unsupervised}. We have also evaluated the performance of the transformer model after the domain adaptation process, calculating the accuracy obtained during anomaly detection. Our best domain-adapted model achieves a global accuracy of $85.45\%$, $83.95\%$ for anomalies and $86.95\%$ for normal signals, while the transformer model achieves a global accuracy of $68.20\%$, $61.59\%$ for anomalies and $74.81\%$ for normal signals, which represents a drop of approximately $17\%$. This accuracy drop shows that our model also provides a better differentiation capability between normal and anomalous signals than the transformer model.

\subsection{Temporal Length study} \label{sec:temporal_study}

Before obtaining our final proposal for the temporal length of $W_I$ in the sample, that is 4s, we studied different values for the temporal length of the input signal to achieve the best possible accuracy without introducing excessively long input samples that require more computation, memory, and inference time. Thus, we employed input signals of 4s, 3s, 2s, and 1s, to predict the following second of unseen signal. Note that, to obtain comparable accuracy results among temporal lengths, we must use the same number of samples in all cases. Thus, we truncate the length of the longest sample, i.e., 4 seconds, input signals generated from both datasets to use only the corresponding number of seconds. Once the samples are generated with the corresponding length, we follow the anomaly detection process described above with a 5-fold cross-validation to obtain the accuracy metric. Table \ref{tab:seconds} summarizes the average accuracy and standard deviation for each temporal length. The best average accuracy was obtained with an input of 4 seconds, which also achieves a more balanced accuracy for normal and anomaly than for the other lengths. In contrast, the worst average accuracy occurs with 3 seconds as input, which is unexpected compared to 2 or 1 seconds as input. A visual analysis of the different behaviors revealed that the 3-second input signals were cut in a critical part of the structure of the ECG, which explained the poorer performance. Finally, we conclude that using the 4-second input is the best option, as it provides a good balance between accuracy and the length of the input signal.

\begin{table}[tbh]
\centering
\scriptsize
\begin{tabular}{cccc}
\hline
\textbf{Input}       & \textbf{Global (\%)} & \textbf{Anomalies (\%)} & \textbf{Normal (\%)} \\ 
\hline
4 seconds & \textbf{84.65 ± 0.56}&\textbf{83.84 ± 2.97}&85.46 ± 2.99\\ 
3 seconds &75.83 ± 0.57&63.95 ± 3.27&\textbf{87.71 ± 3.70}\\ 
2 seconds &83.42 ± 0.84&80.50 ± 5.46&86.34 ± 6.89\\ 
1 seconds &82.13 ± 0.46&80.25 ± 2.81&84.01 ± 2.45\\ 
\hline
\end{tabular}
\caption{\textbf{Temporal Length Study.} 5-fold average accuracy and standard deviation using different input seconds.}
\label{tab:seconds}
\end{table}

\subsection{Understanding what FADE has learned}

In this subsection, our aim is to understand what our system is learning to help doctors who use our proposal. To this end, we conducted two different studies: one visualizing the highest activations and the other analyzing which anomalies are well or poorly predicted by our model in order to understand which classes are confused by our proposal.

\begin{figure*}[tbh]
\centering
    \begin{subfigure}[b]{0.27\textwidth}
    \centering
    \includegraphics[width=\textwidth, height=4cm]{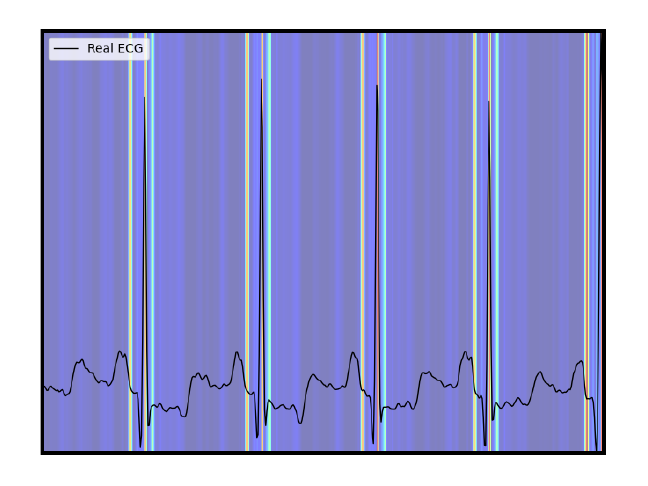}
    \caption{Slow Path\label{fig:slow}}
    \end{subfigure}
\quad
    \begin{subfigure}[b]{0.27\textwidth}
    \centering
    \includegraphics[width=\textwidth,height=4cm]{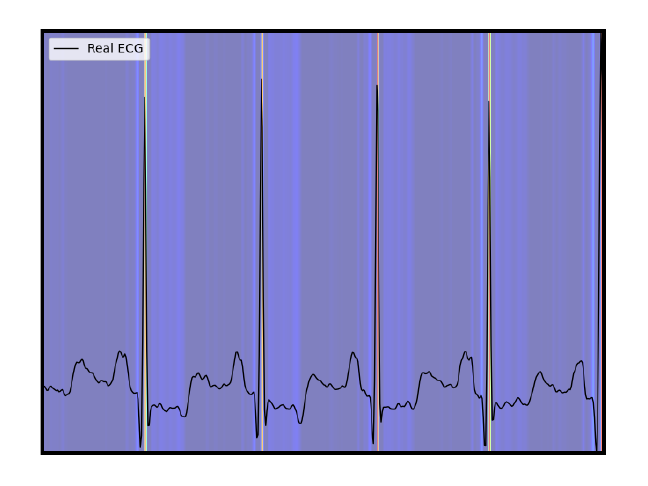}
    \caption{Fast Path\label{fig:fast}}
    \end{subfigure}
    \quad
\begin{subfigure}[b]{0.33\textwidth}
    \centering
    \includegraphics[width=\textwidth, height=3.94cm]{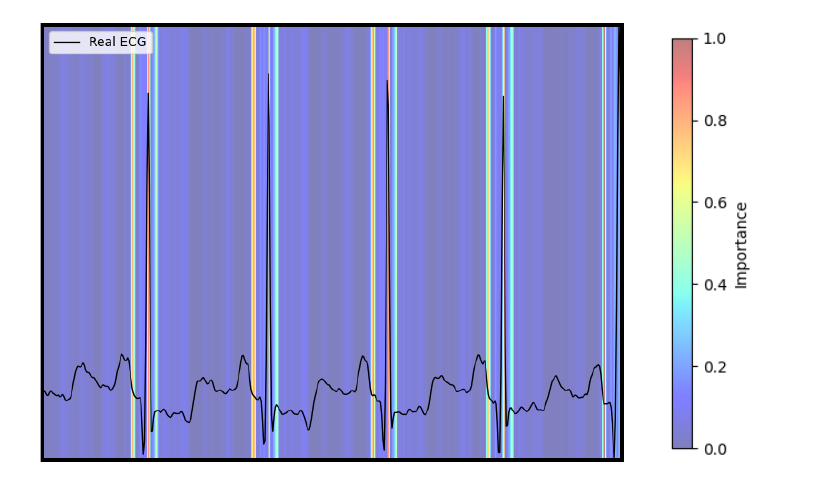}
    \caption{Combination\label{fig:comb}}
    \end{subfigure}
\caption{\textbf{Activation Heatmaps.} Visualization of three heatmaps obtained in the first convolution of Slow Path and Fast Path using an ECG as input (a) First Convolution Slow Path. (b) First Convolution Fast Path. (c) Combination of both paths.}
\label{fig:heatmaps}
\end{figure*}

On the one hand, we have calculated the heat maps over the first convolutional layers in each path of the model (Slow Path and Fast Path) to determine which ECG parts the model focuses on the most. We cannot use Grad-CAM heat maps \cite{selvaraju2017grad} due to the lack of a classification layer. Therefore, we have adapted this method to obtain activation heat maps with our model. First, we calculate the sum of the activations and we compute the gradients. Next, we condense the gradient information by averaging them. Later, the averaged gradients are used to weight the activations. Finally, the heat map is generated by averaging the weighted activations across the feature maps. The results of this procedure can be observed in Figure \ref{fig:heatmaps}. Figure \ref{fig:slow} shows the heat map from the first convolution layer of the Slow Path, Figure \ref{fig:fast} shows the heat map from the first convolution layer of the Fast Path and Figure \ref{fig:comb} represents the combination of the heat maps of both paths. We can conclude that the model focuses mostly on R-peaks: Fast Path exactly on the R-peaks and Slow Path on their surroundings, whereas the rest of the ECG has more or less the same importance for the model. 

On the other hand, we take the best domain-adapted model, that is, the model that achieves a global accuracy of $85.45\%$, an accuracy of $83.95\%$ for anomalies, and $86.95\%$ for normal signals. Then, we calculate the accuracy per anomaly using only the test signals, excluding the subset used for threshold establishment. Note that we do not add a classifier to our model, we made this classification using the threshold that divides normal signals and anomalous signals. The accuracy results for each class can be observed in Table \ref{tab:classes}. The anomalies defined as a sum of two anomalies are due to both anomalies being present during the second of our label signal. From these results, we can conclude that our model has better precision in detecting rhythm anomalies than in detecting beat anomalies, with $97.48\%$ of accuracy versus $79.78\%$ of accuracy. Moreover, we have analyzed the five classes that achieved less than $70\%$ of accuracy. Atrial premature beat, with an accuracy of $50.69\%$, is characterized by an extra heartbeat occurring prematurely, prior to the next expected normal beat. Its morphology is not significantly different from that of normal beats, which likely contributes to the difficulty in distinguishing it from non-anomalous classes. Similarly, the fusion of ventricular and normal beat, which achieved an accuracy of $60.78\%$, is partially formed by a normal beat, which may explain its low precision. Finally, the remaining classes with accuracies below 70\% contain fewer than 12 samples per class, which means that only a few errors will affect significantly the accuracy of the model.

\begin{table*}[tbh]
\scriptsize
\renewcommand{\arraystretch}{1.1}
\setlength{\tabcolsep}{0.3em}
\centering
\begin{tabular}{l|lc}
\hline
\multicolumn{1}{c|}
{\begin{tabular}[l]{@{}l@{}}\textbf{Type of}\\\textbf{Anomaly}\end{tabular}} & \multicolumn{1}{c}{\textbf{Anomaly}} & \multicolumn{1}{c}{\textbf{Accuracy (\%)}} \\ \hline
\multirow{7}{*}{\begin{tabular}[l]{@{}l@{}}\textbf{Rhythm}\\\textbf{Anomaly}\end{tabular}} 
  & Atrial bigeminy & 100 \\ 
  & Atrial fibrillation & 100 \\ 
  & Pre-excitation & 82.35 \\ 
  & Supraventricular tachyarrhythmia & 100 \\ 
  & Ventricular bigeminy & 100 \\ 
  & Ventricular tachycardia & 100 \\ 
  & Ventricular trigeminy & 100 \\ \hline
\multirow{15}{*}{\begin{tabular}[l]{@{}l@{}}\textbf{Beat}\\\textbf{Anomaly}\end{tabular}} 
  & Aberrated atrial premature beat & 77.78\ \\ 
  & Atrial premature beat & 50.69\\ 
  & Atrial premature beat + Atrial escape beat &  66.67\\ 
  & Atrial premature beat + Premature ventricular contraction & 100 \\ 
  & Fusion of paced and normal beat & 92.31\\ 
  & Fusion of paced and normal beat + Premature ventricular contraction & 100 \\ 
  & Fusion of ventricular and normal beat & 60.78\\ 
  & Fusion of ventricular and normal beat + Premature ventricular contraction & 100\\ 
  & Nodal (junctional) escape beat & 0\\ 
  & Premature ventricular contraction &  98.43\\ 
  & Premature ventricular contraction + Aberrated atrial premature beat & 100\\ 
  & Premature ventricular contraction + Atrial premature beat &  100\\ 
  & Premature ventricular contraction + Fusion of paced and normal beat &  100\\ 
  & Right bundle branch block beat & 66.67\\ 
  & Unclassifiable beat & 83.33\\ \hline
\end{tabular}
\caption{\textbf{Anomalies Accuracy.} Each row represents a different anomaly, the anomalies are divided in rhythm and beat anomalies. Note that the anomalies defined as a sum of two anomalies are due to both anomalies being present during the second of our label signal.}
\label{tab:classes}
\end{table*}

\subsection{Ablation Study}
In this subsection we conduct two ablation studies: one to analyze different variations of our proposed Split-MSE loss and another to validate the necessity of using both slow and fast paths in our model.

\subsubsection{Split-MSE Loss Function Variations}
In this subsection, we evaluate the accuracy of our proposed system using variations of the Split-MSE loss function presented previously in Section \ref{sec:loss}. Specifically, we have employed three variations: (i) using only the outer band from Figure \ref{fig:loss}, (ii) using only the inner band from Figure \ref{fig:loss}, and (iii) separating positive and negative values into two distinct bands. The model achieves a global accuracy of $67.11\%$, $68.94\%$ and $66.04\%$ for these variations, respectively. These values differ significantly from the accuracy obtained using our original Split-MSE loss, which reaches a global accuracy of $85.46\%$. Thus, we can conclude that is necessary to guide the model to learn how to generate both, the extreme points (R and S in Figure \ref{fig:loss}) as well as the points closest to 0. The accuracy comparison for anomalies and normal signals using these variations can be observed in the first part of Table \ref{tab:ablation}, as well as the global accuracy.

\subsubsection{Validation of Slow and Fast paths}

In this subsection, we demonstrate the necessity of employing both Slow and Fast paths in our proposed approach. On the one hand, when we implement only the Slow path in combination with the U-Net based decoder, the model achieves a global accuracy of $83.44\%$, with an accuracy of $84.05\%$ for anomalies and $82.82\%$ for normal signals. However, this configuration results in a global accuracy approximately $2\%$ lower than that of our proposed model. On the other hand, employing only the Fast path together with the U-Net based decoder yields a global accuracy of $84.53\%$, with an accuracy of $91.61\%$ for anomalies and $77.46\%$ for normal signals. While the global accuracy is slightly higher than that of the Slow path alone, this model exhibits a significant imbalance between the two classes and still performs worse than our proposed approach. These results highlight the role of integrating both paths for achieving superior and balanced performance. The accuracy comparison for anomalies and normal signals using only one path can be observed in the last part of Table \ref{tab:ablation}, as well as the global accuracy.

\begin{table}[tbh]
\scriptsize
\centering
\begin{tabular}{c|c|ccc}
\hline
\multirow{2}{*}{\textbf{Branch}} & \multirow{2}{*}{\textbf{Loss}} & \multicolumn{3}{c}{\textbf{Accuracy (\%)}} \\
                        &                       & \textbf{Global}    & \textbf{Anomalies}   & \textbf{Normal}   \\ \hline
               \multirow{4}{*}{Both}     & Positive Negative     &   66.04        & 80.80            &     51.28     \\
           & Outer Band            &   67.11        & 76.92            &       57.31   \\
         & Inner Band            &    68.94       & 68.52            &    69.36      \\
        & Split-MSE             &   \textbf{85.45}        &    83.95         &     86.95     \\
 \hline\hline
Slow                    & Split-MSE             &  83.44         &    84.05         &     82.82     \\
Fast                    & Split-MSE             &     84.53      &    91.61         &     77.46     \\ \hline
\end{tabular}
\caption{\textbf{Ablation studies results.} Global accuracy, anomalies accuracy and normal signals accuracy using different variations of MSE loss and employing only the slow path, only the fast path or both.}
\label{tab:ablation}
\end{table}

\section{Discussion}
\label{discussion}

Our proposed approach showed great performance in the early detection of any kind of heart anomaly in ECG signals. In contrast to previous works~\cite{kiyasseh2021clocs,wei2022contrastive,lan2022intra,phan2022multimodality,jin2024self} that only detects a small number of anomalies, mainly related to the frequency, i.e., arrhythmias, our approach is more robust since it is able to detect abnormal heartbeats and arrhythmias. Therefore, the impact of this work is greater than that of other previous works and can be summarized as follows:

\begin{itemize}
 
    \item Reduction of healthcare costs. Automating the analysis of ECG signals can significantly reduce the costs associated with manual examination by healthcare professionals. It frees up valuable time for clinicians, allowing them to focus on more complex cases and other important aspects of patient care.
    \item Remote monitoring and telemedicine. With the rise of wearable health devices and telemedicine, there is an increasing need for automated systems to monitor and analyze ECG data remotely. Automated classification enables continuous, real-time monitoring of patients outside clinical settings, enhancing the ability to manage and respond to heart conditions promptly.
    \item Handling large volumes of data. Hospitals and clinics generate vast amounts of ECG data daily. Manually analyzing this data is time-consuming and prone to human error. Automated classification systems can process and analyze large datasets quickly and consistently, ensuring that no critical anomalies are missed. 
\end{itemize}

However, our approach also has some limitations that we will study in the future to improve them. The main limitation is the requirement of $W_{I}$ normal seconds of the signal to forecast $W_{L}$ future seconds. As observed in the MIT-BIH Arrhythmia database, some patients can have abnormal signals all the time, making it impossible to use our system. A possible solution would be to reduce the number of seconds of the input sample as studied in Section~\ref{sec:temporal_study} although, depending on the anomaly, it could not be enough, so it requires an in-depth study to address this limitation. The other main limitation is the impossibility of classifying the type of anomaly. Although our system has an incredible detection capacity, these detections are generic and limit the impact of the system on determining the most critical anomalies for health.

\section{Conclusion}
\label{conclusions}
In this work, we have introduced a novel system for the detection of ECG anomalies that minimizes the reliance on manual interpretation, thus improving diagnostic efficiency and adaptability in different clinical environments, for example, different patients or sensors. Our proposed deep learning system is trained in a self-supervised manner using a novel morphological-inspired loss function to forecast normal ECG signals, eliminating the need for extensive labeled datasets. Using a distance function to compare predicted ECG signals with actual sensor data, our method efficiently detects cardiac anomalies. The proposed system can be adapted to new contexts through domain adaptation techniques, extending its practical applicability. Experimental evaluations using the MIT-BIH NSR and MIT-BIH Arrhythmia datasets show that our system achieves an average accuracy of 83.84\% in detecting anomalies and 85.46\% in correctly classifying normal ECG signals.
In future work, we plan to optimize our proposed system for deployment in wearable devices. Additionally, we aim to extend its functionality to detect anomalies and classify the type of detected anomaly.

\section*{Acknowledgments}

This work has been supported by the Junta de Andaluc\'ia of Spain (UMA20-FEDERJA-059) and the Ministry of Education of Spain (PID2022-136575OB-I00). Moreover, this research was supported in part by the Swiss National Science Foundation Sinergia grant 193813: "PEDESITE -
Personalized Detection of Epileptic Seizure in the Internet of Things (IoT) Era", and the Wyss Center for Bio and Neuro Engineering: Lighthouse Noninvasive Neuromodulation of Subcortical Structures.

\bibliographystyle{elsarticle-num}
\bibliography{refs}


\end{document}